\documentclass[a4paper,twoside]{article}

\usepackage[utf8]{inputenc} % allow utf-8 input
\usepackage[T1]{fontenc}    % use 8-bit T1 fonts
\usepackage[english]{babel}
\usepackage{hyperref}       % hyperlinks
\usepackage{booktabs}       % professional-quality tables
\usepackage{caption}
\usepackage{subcaption}
\usepackage{siunitx}
\usepackage{rotating}
\usepackage{epsfig}
\usepackage{calc}
\usepackage{amssymb}
\usepackage{amstext}
\usepackage{amsmath}
\usepackage{amsthm}
\usepackage{multicol}
\usepackage{pslatex}
\usepackage{apalike}
\usepackage{SCITEPRESS}     % Please add other packages that you may need BEFORE the SCITEPRESS.sty package.

\newsavebox\CBox
\def\textBF#1{\sbox\CBox{#1}\resizebox{\wd\CBox}{\ht\CBox}{\textbf{#1}}}

\makeatletter
\let\mySection\section
\renewcommand{\section}{\@ifstar{\@secb}{\@seci}}
\newcommand{\@secb}[1]{\mySection*{\uppercase{#1}}}
\newcommand{\@seci}[1]{\mySection{\uppercase{#1}}}
\makeatother

\newcommand{\myIndent}{\noindent}

\begin{document}

%\title{OmniDetector: With Neural Networks to Bounding Boxes}
%\title{A comparison of non-maxima supression approaches for objects from different perspective views generated from fish-eye images}
\title{Improved Person Detection on Omnidirectional Images with Non-maxima Suppression}

\author{
    \authorname{
      Roman Seidel,
      André Apitzsch,
      Gangolf Hirtz}
    \affiliation{Department of Information Technology,
      Chemnitz University of Technology,
      Chemnitz, Germany}
    \email{\{roman.seidel, andre.apitzsch, g.hirtz\}@etit.tu-chemnitz.de}
}

\keywords{Ambient Assisted Living, Convolutional Neural Networks, Object Detection, Non-maxima Suppression, Omnidirectional Images}

\abstract{
% We propose bounding box based ground truth on omnidirectional images, an accurate method to generate minimal enclosing rectangles of persons.
We propose a person detector on omnidirectional images, an accurate method to generate minimal enclosing rectangles of persons.
The basic idea is to adapt the qualitative detection performance of a convolutional neural network based method, namely YOLOv2 to fish-eye images.
The design of our approach picks up the idea of a state-of-the-art object detector and highly overlapping areas of images with their regions of interests.
This overlap reduces the number of false negatives.
Based on the raw bounding boxes of the detector we fine-tuned overlapping bounding boxes by three approaches: non-maximum suppression, soft non-maximum suppression and soft non-maximum suppression with Gaussian smoothing.
The evaluation was done on the PIROPO database and an own annotated Flat dataset, supplemented with bounding boxes on omnidirectional images.
We achieve an average precision of 64.4\,\% with YOLOv2 for the class person on PIROPO and 77.6\,\% on Flat.
For this purpose we fine-tuned the soft non-maximum suppression with Gaussian smoothing.
%The soft non-maximum supression with Gaussian smoothing leads to an improvement of 41\,\% of F1-measure with respect to the public available PIROPO database with omnidirectional images. 
}

\onecolumn \maketitle \normalsize \vfill

\section{Introduction}
\label{sec:intro}
\myIndent
Convolutional neural networks (CNNs) were treaded for several tasks in computer vision in the recent years.
Finding objects in images (i.e.\ object detection) belongs to these tasks.
A main requirement for the detection of objects in images for current CNNs are accurate real-world training data.
%Modeling synthetic data close to real-world data with respect to texture and illumination is a challenge.
In this paper we propose a method to detect objects in fish-eye images of indoor scenes using a state-of-the-art object detector.
% In contrast to ground truth datasets on perspective images, which are widely available, accurate labeled image data from omnidirectional sensors are inaccessible.

The object detection in indoor scenes with a limited number of image sensors can be reached with images from omnidirectional cameras.
These cameras are suited for capturing one room with a single sensor due to a field of view of about \ang{180}.
Our goal is to detect objects in indoor scenes in omnidirectional data with a detector trained on perspective images.

Beside our application, the field of active assisted living, the detection of objects in omnidirectional image data can be used in mobile robots and in the field of autonomous driving.

The remainder of this paper is structured as follows:
Section \ref{sec:related_work} presents previous research activities in object detection.
%in transformation of images from omnidirectional to perspective camera model and 
Section \ref{sec:obj_detection} illustrates the working principle of a neural network based object detector.
Section \ref{sec:virtual_cameras} explains how our virtual cameras are generated.
Section \ref{sec:nms} shows the theoretical background for different variants of non-maximum suppression (NMS). 
Section \ref{sec:experiments} describes our experiments for the generation of bounding boxes and the evaluation of our results on common error metrics.
Section \ref{sec:conclu} summarizes the paper's content, concludes our observations and gives ideas for future work.
The results of our work, the image data and the evaluation of the results, can be found at \url{https://gitlab.com/auxilia/omnidetector}.

\section{Related work}
\label{sec:related_work}
\myIndent
%\subsection{Omnidirectional to Perspective Transformation}
% A way to transform images from omnidirectional to perspective camera geometry was suggested by \cite{FindeisenMHAH13}.
%\subsection{Object Detection on Perspective Images}
State-of-the-art object detectors predict bounding boxes on perspective images over several classes. 
A region-based, fully connected convolutional network for accurate and efficient object detection is R-FCN \cite{Dai2016RFCN}.
As a standard practice, the results of the detector based on ResNet-101 architecture \cite{He_resnet_15} are post-processed with non-maximum suppression (NMS) using a threshold of 0.3 to the intersection over union (IoU) \cite{GirshickDDM13}.
The single shot multi-box detector (SSD) by \cite{Liu2016} provides an improvement of the network architecture by adding a backend extra feature layer on top of VGGNet-16
combined with the idea to use predictions from multiple feature maps with different resolutions which handles objects with various sizes.
The SSD leads to competitive results on common object detection benchmark datasets, namely MS COCO \cite{ms_coco}, ImageNet \cite{Russakovsky2015} and PASCAL VOC \cite{Everingham2010}.
The approach we follow is YOLOv2 \cite{RedmonF2017}.
It produces significant improvements to increase mean average precision (mAP) through variable size of models,
multi-scale training and a joint training to predict detections for object classes without labeled detection data.
%\subsection{Ground Truth generation}

% State of the art ground truth datasets can be divided into two categories.
% Real world datasets with manually or semi-automatically created labels and synthetically generated ground truth.
Our application is object detection, so we concentrate on datasets where labels are minimally enclosing rectangles (bounding boxes).
Common real world benchmark datasets with labeled objects on perspective images are presented by \cite{Everingham2010}, \cite{openimages}, \cite{li_webvision}, \cite{Russakovsky2015}.
% Ground truth generation with synthetic human models and perspective background is presented in SURREAL (Synthetic hUmans foR REAL tasks, \cite{Varol_2017_CVPR}).
%The idea behind SURREAL is the generation of synthetic objects for training object detectors with image quality close to real world images, rather in terms of features during the detector's training \cite{EPFL-ARTICLE-204712}.
%\citet{flownet} study the generation of synthetic ground truth for optical flow using rendered images from blender video sequences.
Omnidirectional images with multiple sequences in two different indoor rooms were created in the work of \cite{piropo2016}.
% While bounding boxes are the common notation for generating ground truth for object detection, \cite{piropo2016} concentrates on annotations as single points on the head of persons.
A direct approach for detecting objects in omnidirectional images without CNNs was shown in the work of \cite{cinaroglu2014direct}.
The classical HoG features and training a SVM to detect humans in a transformed INRIA dataset leads to competitive results in recall and precision.

A novel model named Past-Future Memory Network (PFMN) was proposed by \cite{lee2018memory} on \ang{360} videos.
One of the main contributions of \cite{lee2018memory} is to learn the correlation between input data from the past and future.

In contrast to our work, the authors of Spherical CNN \cite{2018spherical} modify the architecture  of ResNet.
Their goal is to build a collection of spherical layers which are rotation-equivariant and expressive.
% With the application of 3D model recognition the invariance against rotation is reached through a generalized Fast Fourier Transform.
% The input of the network is a 3D model which is rendered from multiple views in a second stage.

\section{Object detection}
\label{sec:obj_detection}
\myIndent
Based on an excellent mAP of 73.4\% (10 classes, VOC2007test) and an average precision (AP) of 81.3\% (VOC2007test) for the class person, we use the \textit{You Only Look Once (YOLO)} \cite{RedmonDGF2016} approach in its second version called YOLOv2 \cite{RedmonF2017}.
To detect objects in input images YOLOv2 offers a good compromise between detection accuracy and speed.
%training			double citation of imagenet (related work) allowed / common?
The model is trained on ImageNet \cite{Russakovsky2015} and the COCO dataset \cite{ms_coco}.
The approach outperforms state-of-the-art methods like Faster R-CNN \cite{NIPS2015_5638} with ResNet \cite{He_resnet_15} and SSD \cite{Liu2016}, which still runs significantly faster.
%architecture
YOLOv2 predicts the corners of bounding boxes directly with the help of fully connected layers which are added on top of the convolutional feature extractor.
Additional changes on the network architecture are the elimination of pooling layers to obtain a higher resolution output by the convolutional layers in the network.
% one pooling less leads to higher resoluted output!!
The input data size of the network is shrinked to operate on \(416 \times 416\) input images instead of \(448 \times 448\).
%advantages /test
For the prediction of bounding boxes in YOLOv2 the fully connected layers are replaced by anchor boxes.
To counteract the effect to detect objects with a fixed size, a special feature during the training is the random selection of input size of the model, which changes every 10 batches.
The smallest input is \(320 \times 320\) and the largest \(608 \times 608\).

\section{Creating virtual views from an omnidirectional image}
\label{sec:virtual_cameras}
\myIndent
In this chapter we describe the transformation for generating virtual perspective views from omnidirectional image data based on \cite{FindeisenMHAH13}.
We assume, that the omnidirectional camera is calibrated both intrinsically and extrinsically.

The camera model describes how the coordinates of a 3D scene point are transformed into the coordinates of a 2D image point.
We concentrate on the central camera model, i.e.\ all light rays, originating from the scene points, travel through a single point in space, called the single effective viewpoint.
For the transformation between the omnidirectional and the perspective images a mathematical description is necessary for both camera models.

\subsection{Perspective Camera Model}
\label{perspective_camera_model}

The perspective camera model uses the pinhole camera model as an approximation.
The perspective projection of the spatial coordinates given in the camera coordinate system is stated \(\mathbf{x}_{cam} = (x_{cam}, y_{cam}, z_{cam})^T\) and in normalized image coordinates \(\mathbf{x}_{norm} = (x_{norm}, y_{norm}, 1)^T\).
After applying an affine transformation it is possible to get pixel coordinates \(\mathbf{x} = (x_{img}, y_{img})^T\).
For the linear mapping between the source and target camera model we use \textit{homogeneous coordinates}, denoted as \(\mathbf{\tilde x} = (x, y, 1)^T\).
The relation between \(\mathbf{x}_{norm}\) and \(\mathbf{\tilde x}\) is given by
\begin{equation}
    \mathbf{\tilde x} = \mathbf{K} \cdot \mathbf{x}_{norm}
    \label{eq:central_projection}
\end{equation} 
where \(\mathbf{K}\) is the upper-triangular calibration matrix containing the camera intrinsic:
\begin{equation}
    \mathbf{K} =
    \begin{bmatrix}
    f_x & s_{\alpha}& c_x \\
    0   & f_y       & c_y \\
    0   & 0         & 1
    \end{bmatrix}.
\label{eq:K_mat}
\end{equation}
As shown in \eqref{eq:K_mat} the five intrinsic parameters of a pinhole camera are
the scale factors in x- and y-direction \((f_x, f_y)\), the skewness factor \(s_{\alpha}\) and the principle point of the image \((c_x, c_y)\).

In general a scene point is modeled in a world coordinate system, which is different from the camera coordinate system (\(\mathbf{x}_{cam})\).
The orientation between these coordinate systems consists of two parts, namely a rotation \(\mathbf{R}\) and a translation \(\mathbf{t}\) (or equivalent \(\mathbf{C}=-\mathbf{R}^{-1}\cdot\mathbf{t}\), where \(\mathbf{C}\) is the camera center).

The relationship between the scene point in the world coordinate system \(\mathbf{\tilde X} = (X, Y, Z, 1)^T\) and an image point in the image coordinate system \(\mathbf{\tilde x}\) is given by
\begin{equation}
    \mathbf{\tilde x} = \mathbf{P} \cdot \mathbf{\tilde X}
\label{eq:projection_intandext}
\end{equation}
where \(\mathbf{P}\) is a homogeneous \(3 \times 4\) matrix, called the \textit{camera projection matrix} \cite{Hartley2004}.
The matrix \(\mathbf{P}\) contains the parameters of the extrinsic and intrinsic calibration with
\begin{equation}
    \mathbf{P} = \mathbf{K}[\mathbf{R}\vert\mathbf{t}].
    \label{eq:P_matrix}
\end{equation} 
There are several approaches to extend the camera model defined above with a description of lens imperfections.
As long as our target virtual camera is perfectly perspective and free of lens distortions, we do not discuss this issue.

\subsection{Omnidirectional to Perspective Image Mapping}
\label{virtual_cameras}
% -->hier weiter (2018-01-11)
Because it is mathematically impossible to transform the whole omnidirectional image into one perspective image, we transform a region of 2D image points from the omnidirectional into the perspective view.
We determine the perspective images through \textit{n} virtual perspective cameras \(\textit{Cam\textsubscript{0}}\), \(\textit{Cam\textsubscript{1}}\), \ldots, \(\textit{Cam\textsubscript{n}}\),
which are described by their extrinsic parameters \(\textbf{R}\) and \(\textbf{t}\) (6 degrees of freedom (DOF)) and intrinsic parameters \(\mathbf{K}\) (5 DOF).
Instead of determining the parameters of the perspective camera through a calibration, we model the virtual camera and determine the extrinsic (\(\textbf{R}\) and \(\textbf{t}\)) empirically.

To create the virtual perspective views we change the extrinsic camera parameter \(\textbf{R}\) through the variation of the angles through the rotation about the axes x, y and z represented by their Euler angles.
To be more specific, we rotate about the x-axis and z-axis.

The extrinsic calibration parameters of the omnidirectional camera form the world reference with respect to the virtual perspective cameras.
As \(\textbf{K}\) contains the scale factors in the horizontal and vertical directions (\(f_x, f_y\)), \(\textbf{K}\) determines the field of view (FOV) of the target images.
For perspective images with a resolution of \(2c_x \times 2c_y\) the horizontal and vertical FOVs are:
\begin{equation}
	\begin{split}
		&FOV_h = 2 \arctan\left(\frac{c_x}{2f_x}\right) \text{ and } \\
		&FOV_v = 2 \arctan\left(\frac{c_y}{2f_y}\right),\text{ respectively}.
	\end{split}
	\label{fov_hv}
\end{equation}
Equation \eqref{fov_hv} allows us to define the FOV of the perspective camera and to build at least one virtual perspective camera,
which is able to generate perspective images, from the omnidirectional camera.
Derived from the horizontal FOV and vertical FOV  we determine the diagonal FOV (\(FOV_d\)) with:
\begin{equation}
		FOV_d = 2 \arctan\left(\frac{c}{2f}\right)
	\label{fov_d}
\end{equation}
where:
\begin{equation}
	\begin{split}
		c = \sqrt{c_x^2 + c_y^2} \text{\quad and \quad}
		f = \sqrt{f_x^2 + f_y^2}.
	\end{split}
	\label{c_f}
\end{equation}
To come to a common FOV of a usual perspective camera we choose the focal length and the diagonal image size with respect to the sensor to be equal.
This leads to a simplification of \eqref{fov_d} with:
\begin{equation}
    FOV_d = 2 \arctan\left(\frac{1}{2}\right).
	\label{fov_d_simple}
\end{equation}
The simplification leads to a diagonal FOV of about \ang{53.13} and allows us to choose \(c\) and \(f\) free, as long as they are equal.

\section{Non-maximum suppression}
\label{sec:nms}
\myIndent
Our goal is to find the most likely position of the minimal enclosing rectangle of the object.
Therefore we disable the two final steps of YOLOv2 occurring at the last layers of the network.
First, the reduction of the number of bounding boxes based on their confidence.
Second, the union of multiple bounding boxes of one particular object through soft non-maximum suppression (Soft-NMS).

In general, the NMS is necessary due to highly overlapping areas of perspective images after the transformation to omnidirectional images. 
To receive the raw detections of YOLOv2 with confidences between 0 and 1, we set the confidence threshold equal to zero.
To group the resulting bounding boxes, one suitable measurement is the intersection over union (IoU).
The IoU for two boxes \(A\) and \(B\) is defined by the Jaccard index as:
\begin{equation}
  IoU(A,B) = \frac{A \cap B}{A \cup B}.
  \label{eq:iou}
\end{equation}
Our next step for the refinement of the back-projected bounding boxes is applying Soft-NMS inspired by \cite{DBLP:conf/iccv/BodlaSCD17}.
In this approach Soft-NMS is used to separate bounding boxes to distinguish between different objects that are close to each other and to prune multiple detections for one unambiguous object, back projected from highly overlapped perspective views.
Bounding boxes which are close together and fulfill the IoU \textgreater{} 0.5 are considered as a unique region of interest (RoI) proposal for each object.
To update the confidences of the bounding boxes, in the NMS the pruning step can be formulated as a rescoring function:
\begin{equation}
    s_{i} = 
    \begin{cases}
	s_{i}, & IoU(M,b_{i}) < N_{t}, \\
	0, & IoU(M,b_{i}) \ge N_{t}.
    \end{cases}
    \label{eq:nms}
\end{equation}
Where \(b_{i}\) is a bounding box with score \(s_i\) of the detector and \(M\) is the detection box with maximum score.
The parameter \(N_{t}\) describes the NMS threshold, which removes boxes from a list of detections with certain scores, as long as the \(IoU(M,b_{i})\) is greater than or equal to the NMS threshold.
The result of \eqref{eq:nms} is a confidence score between zero and one, which is used to decide what is kept or removed in the neighborhood of \(M\).

The Soft-NMS approach is able to weight the score of boxes \(b_{i}\) in the neighborhood of \(M\).
\begin{equation}
    s_{i} = 
    \begin{cases}
	s_{i}, & IoU(M,b_{i}) < N_{t}, \\
	s(1- IoU(M,b_{i})), & IoU(M,b_{i}) \ge N_{t}.
    \end{cases}
    \label{eq:soft_nms}
\end{equation}
Equation \eqref{eq:soft_nms} describes the rescoring function for the Soft-NMS.
The goal is to decay the scores above a threshold \(N_{t}\) modeled with a linear function.
The scores of the bounding boxes from the detection with a higher overlap with \(M\) have a stronger potential of being false positives.
As a result we get a rating of the bounding boxes \(b_{i}\) with respect to \(M\) without changing the number of boxes.
With an increasing overlap between detection boxes and \(M\) the penalty increases.
At a low overlapping area between \(b_{i}\) and \(M\) the scores will be not affected.
To penalize \(b_{i}\) stronger if the \(IoU\) becomes close to one, the pruning step can also be modeled as a Gaussian penalty function:
\begin{equation}
    s_{i} = s_{i} \cdot e^{-\frac{IoU(M,b_{i})^{2}}{\sigma}}\text{, } \forall b_{i} \in \mathcal{B} \setminus \mathcal{D}, 
    \label{eq:s_nms_gaussian}
\end{equation}
where \(\mathcal{B}\) is the set of back-projected raw detections of YOLOv2 and \(\mathcal{D}\) is a growing set of final detections.

\section{Experimental Results}
\label{sec:experiments}
\myIndent
We evaluate our approach on two datasets, that are single images from an omnidirectional camera of an indoor scene.
To qualitatively evaluate our detection results we use a labeled image dataset from omnidirectional camera geometry, namely the PIROPO database (People in Indoor ROoms with Perspective and Omnidirectional cameras).
The input images have a resolution of \(600 \times 600\) pixels, are undistorted and captured with a ceiling-mounted omnidirectional camera.
The image data contain point labels on the head of persons.
To compare the results of the detection with respect to the ground truth, we manually create bounding box ground truth for the class person in 638 images.
The subset of the labeled data of the PIROPO database is available on the website mentioned in Section \ref{sec:intro}.
Subsequently, we create a new dataset with multiple persons moving in a room, that we call \textit{Flat}. The images of this dataset have a resolution of \(1680 \times 1680\) pixels.

We assume, that our start point is an image from a virtual perspective camera.
The creation of virtual perspective views from omnidirectional images is described in Section \ref{virtual_cameras}.
We made several experiments to validate the deterministic behavior of YOLOv2 by choosing different confidence values for the detection boxes.
While the location of the bounding box in the image is variable through reproducible attempts, for generating the results we keep the confidence value of the detector \((0.8)\) constant for true positive detections.

The way, we create the perspective images from our omnidirectional image data, is described as follows:
We vary both the rotation around the x-axis and z-axis.
The rotation around the z-axis corresponds to the azimuth of the omnidirectional camera model.
Rotating around the x-axis matches to the elevation of the omnidirectional camera model.
The elevation is changed from \(0.0\) to \(0.9\) with a step size of \(0.3\).
We choose the four different perspective views to avoid the black image proportion at the boundaries of the omnidirectional image, which does not contain additional information.
The azimuth is changed from \(-3.14\) to \(3.14\) with a step size of \(0.2\),
for covering the whole room with perspective views.
% The high amount of the overlapping areas of the perspective images is motivated through two aspects.
% First, the reliable detection of an object on the perspective image through a detection rate that is smaller than the ground truth.
% Second, to avoid false positive detections, we choose a high overlap of the perspective images to not miss objects on the perspective image boundaries. 

As an additional constraint, we assume in our configuration (camera's mounting height with respect to the room size) that the person fits in one perspective image.
After the calculation of the detection results in the perspective images, we transform these detections to the omnidirectional source image.

The use of a look-up-table (LUT) for back projecting the perspective images to the omnidirectional image leads to their original position of the source image in the target image.
Additionally, the corners of the bounding boxes are also transformed with the help of the LUT.
Through the back transformation of the bounding box corners the new boxes become larger.

\subsection{Bounding Box Refinement}
% 1: extract/exclude nms from yolo
For the grouping of bounding boxes based on their confidences the YOLOv2 object detector has an included NMS, as described in Section \ref{sec:nms}.
If the IoU is higher than a threshold \(N_t\), then multiple boxes of an object are merged.
% We eliminate the included NMS in the YOLOv2 and do it manually as a modular postprocessing step to check the deterministic behavior of the algorithm.
With the help of a small test set, we evaluate YOLOv2's confidence both with the internal NMS and external NMS, which produces the same confidence values with equal thresholds.
%2: explain the use three aproaches for bb - refinement
To refine multiple bounding boxes projected from the perspective views in the omnidirectional image we use three variants of NMS.

\textbf{NMS}
First, we apply the classical NMS (see \eqref{eq:nms}) to reduce bounding boxes with a predefined overlap threshold \(N_t\).
% name threshold - values and reference to the pr-curves
We vary the overlap threshold \(N_t\) from \(0.0\) to \(1.0\) with a step size of \(0.1\).

\textbf{Soft-NMS}
Second, the use of Soft-NMS (see \eqref{eq:soft_nms}). The advantage of Soft-NMS is penalizing detection boxes \(b_i\) with a higher overlap to \(M\) as long as they are false positives.
Based on modeling the overlap of \(b_i\) to \(M\) as a linear function the threshold \(N_t\) controls the detection scores.
To be more precise, the detection boxes with high distance to \(M\) are not influenced through the function in \eqref{eq:soft_nms}.
The boxes that are close together allocate a high penalty.
% name threshold-values and reference to the pr-curves

\begin{figure*}[ht!]
\centering
    \begin{sideways}
        \hskip 0.05\textwidth
        \small \(C_{t} = 0.3\)
    \end{sideways}
    \begin{subfigure}[b][0.195\textwidth]{.19\linewidth}
        \caption{\(\sigma=0.1\)}
        \fbox{\includegraphics[width=\textwidth]{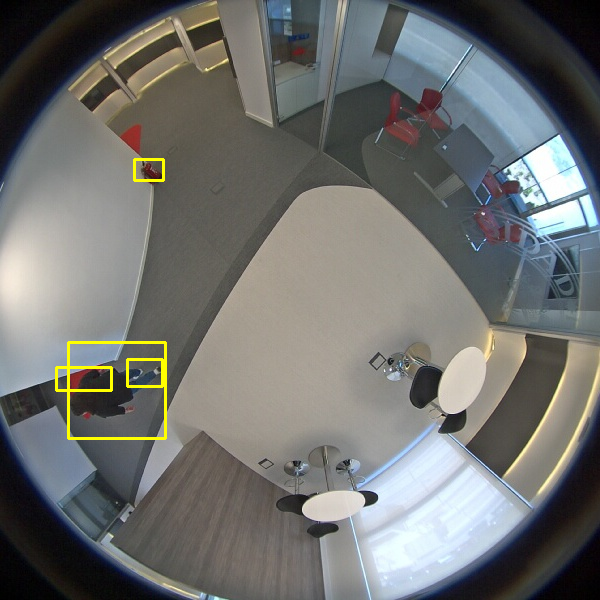}}
    \end{subfigure}
    \begin{subfigure}[b]{.19\linewidth}
        \caption{\(\sigma=0.3\)}
        \fbox{\includegraphics[width=\textwidth]{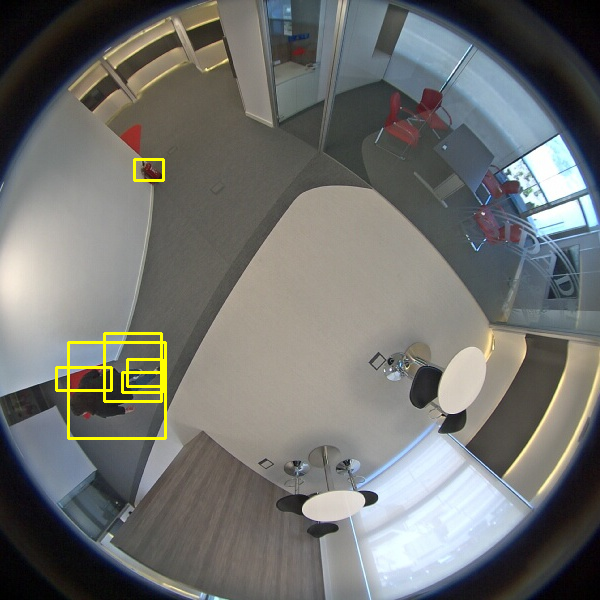}}
    \end{subfigure}
    \begin{subfigure}[b]{.19\linewidth}
        \caption{\(\sigma=0.5\)}
        \fbox{\includegraphics[width=\textwidth]{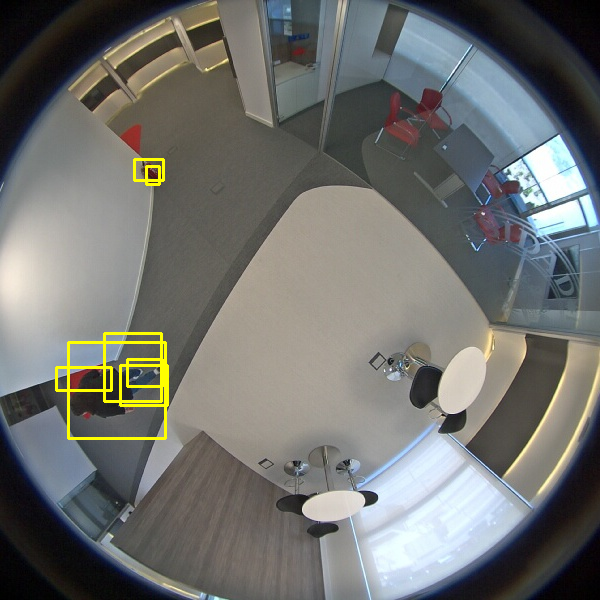}}
    \end{subfigure}
    \begin{subfigure}[b]{.19\linewidth}
        \caption{\(\sigma=0.7\)}
        \fbox{\includegraphics[width=\textwidth]{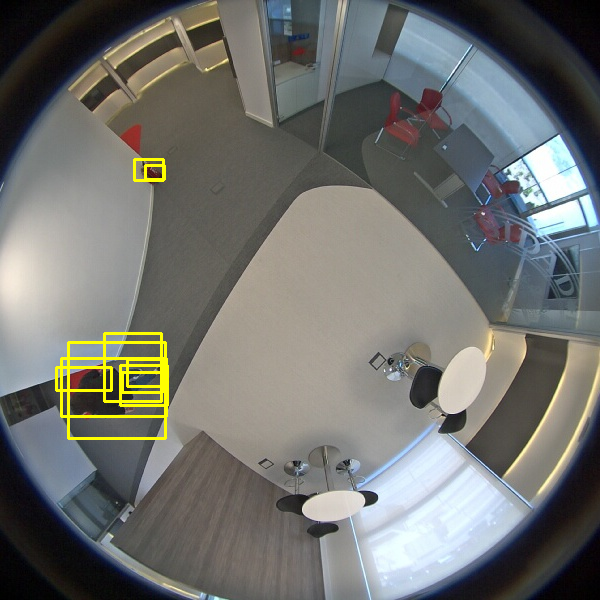}}
    \end{subfigure}
    \begin{subfigure}[b]{.19\linewidth}
        \caption{\(\sigma=0.9\)}
        \fbox{\includegraphics[width=\textwidth]{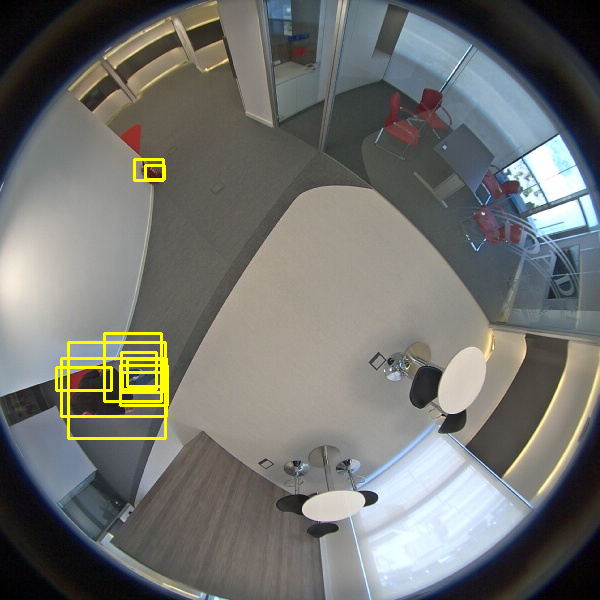}}
    \end{subfigure}
%     \begin{sideways}
%         \hskip 0.05\textwidth
%         \small \(C_{t} = 0.5\)
%     \end{sideways}
%     \begin{subfigure}[b][0.195\textwidth]{.19\linewidth}\fbox{\includegraphics[width=\textwidth]{./figures/result_matrix/0501.png}}\end{subfigure}
%     \begin{subfigure}[b]{.19\linewidth}\fbox{\includegraphics[width=\textwidth]{./figures/result_matrix/0503.png}}\end{subfigure}
%     \begin{subfigure}[b]{.19\linewidth}\fbox{\includegraphics[width=\textwidth]{./figures/result_matrix/0505.png}}\end{subfigure}
%     \begin{subfigure}[b]{.19\linewidth}\fbox{\includegraphics[width=\textwidth]{./figures/result_matrix/0507.png}}\end{subfigure}
%     \begin{subfigure}[b]{.19\linewidth}\fbox{\includegraphics[width=\textwidth]{./figures/result_matrix/0509.png}}\end{subfigure}
%     \begin{sideways}
%         \hskip 0.05\textwidth
%         \small \(C_{t} = 0.7\)
%     \end{sideways}
%     \begin{subfigure}[b][0.195\textwidth]{.19\linewidth}\fbox{\includegraphics[width=\textwidth]{./figures/result_matrix/0701.png}}\end{subfigure}
%     \begin{subfigure}[b]{.19\linewidth}\fbox{\includegraphics[width=\textwidth]{./figures/result_matrix/0703.png}}\end{subfigure}
%     \begin{subfigure}[b]{.19\linewidth}\fbox{\includegraphics[width=\textwidth]{./figures/result_matrix/0705.png}}\end{subfigure}
%     \begin{subfigure}[b]{.19\linewidth}\fbox{\includegraphics[width=\textwidth]{./figures/result_matrix/0707.png}}\end{subfigure}
%     \begin{subfigure}[b]{.19\linewidth}\fbox{\includegraphics[width=\textwidth]{./figures/result_matrix/0709.png}}\end{subfigure}
    \begin{sideways}
        \hskip 0.05\textwidth
        \small \(C_{t} = 0.8\)
    \end{sideways}
    \begin{subfigure}[b][0.195\textwidth]{.19\linewidth}
        \fbox{\includegraphics[width=\textwidth]{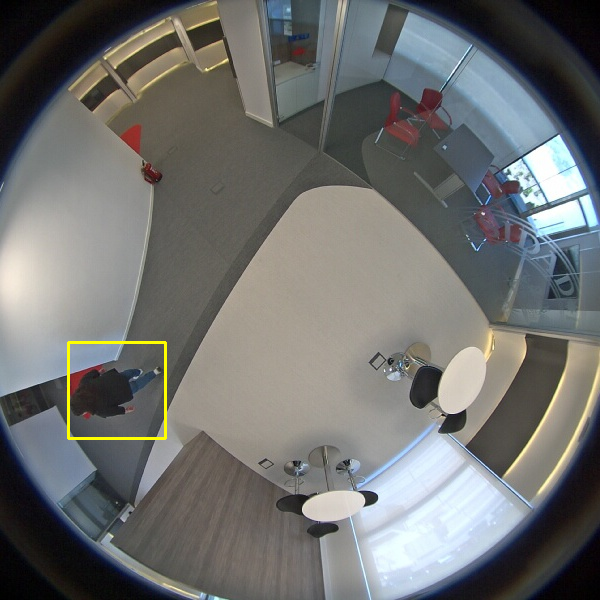}}
    \end{subfigure}
    \begin{subfigure}[b]{.19\linewidth}
        \fbox{\includegraphics[width=\textwidth]{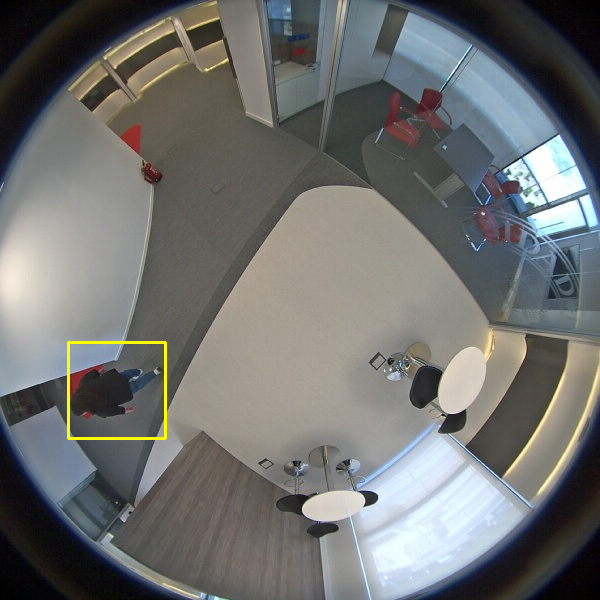}}
    \end{subfigure}
    \begin{subfigure}[b]{.19\linewidth}
        \fbox{\includegraphics[width=\textwidth]{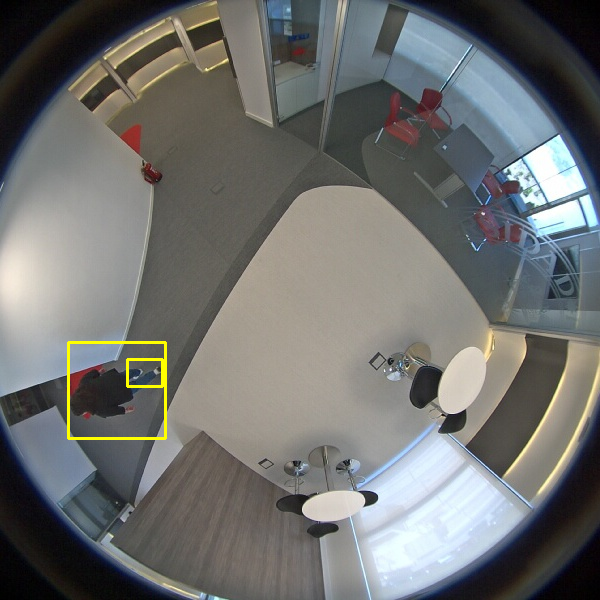}}
    \end{subfigure}
    \begin{subfigure}[b]{.19\linewidth}
        \fbox{\includegraphics[width=\textwidth]{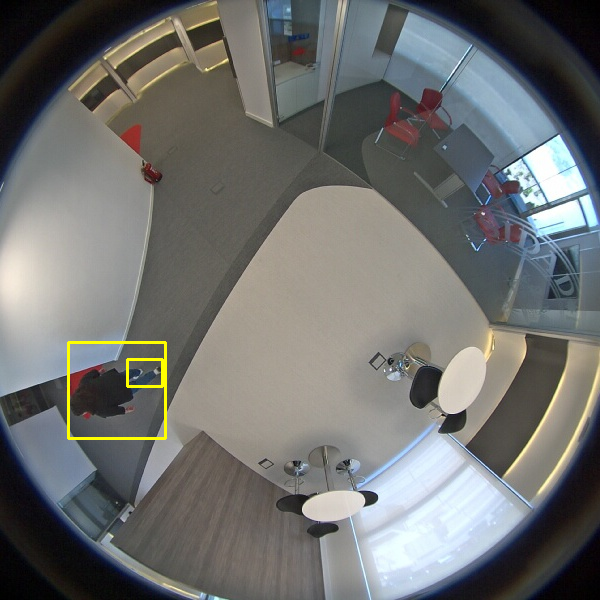}}
    \end{subfigure}
    \begin{subfigure}[b]{.19\linewidth}
        \fbox{\includegraphics[width=\textwidth]{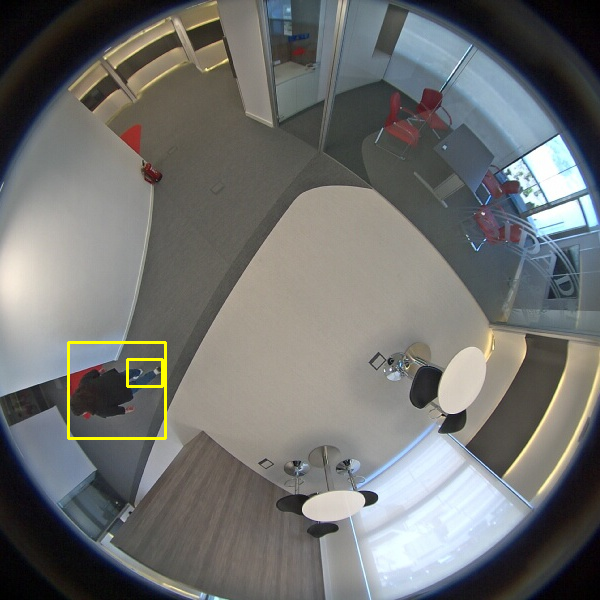}}
    \end{subfigure}
    \caption{Visualization of results for Soft-NMS with the Gaussian penalty function on an omnidirectional image from the PIROPO database. The columns are varied over different \(\sigma\) of the Gaussian penalty function.
    In each row we change the confidence threshold \(C_{t}\) of the Soft-NMS. See text for details.\label{fig:results_snms_gauss}}
\end{figure*}

\textbf{Soft-NMS with Gaussian smoothing}
Third, to retort the problem of abrupt changes to the ranked list of detections, we consider the Gaussian penalty function as shown in \eqref{eq:s_nms_gaussian}.
The Gaussian penalty function is a continuous exponential function, which delivers no penalty in case of no overlap of the boxes and a high penalty at highly overlapped boxes.
The update was done iteratively to all scores of the remaining detection boxes.
Starting from the detectors raw data, we vary the confidence threshold \(C_{t}\) with the values \(0.3\), \(0.5\), \(0.7\) and \(0.8\)
and the Gaussian smoothing factor \(\sigma\) with the values \(0.1\), \(0.3\), \(0.5\), \(0.7\) and \(0.9\).
The corresponding results in Figure~\ref{fig:results_snms_gauss} show a single image from the PIROPO database with the below mentioned variations of thresholds in the rows and columns, respectively.
\setlength{\fboxsep}{0pt}
An effect, which is easily visible is the changing number of bounding boxes in the images.
In the top right corner of the matrix (\(\sigma = 0.9\) and \(C_{t} = 0.3\)) the number of boxes for possible candidates of true positives is high.
The opposite effect, less number of true positives with a high accuracy is observable in the bottom left corner of Figure~\ref{fig:results_snms_gauss} (values of \(\sigma = 0.1\) or \(\sigma = 0.3\) and \(C_{t} = 0.8\)).
Using \(\sigma\) for the steering of the smoothness of the merging of the bounding boxes makes the effects explainable.
The higher we select \(\sigma\), the closer comes the exponential function in \eqref{eq:s_nms_gaussian} to 1.
Is the exponential function close to or equal to 1, the number of boxes does not change.
With the knowledge, that the exponential function cannot become zero, the smaller we set \(\sigma\), the smaller is the number of the bounding boxes in the final set \(\mathcal{D}\).
The Gaussian smoothing function in the Soft-NMS delivers the best results, compared to the other variants of NMS.

\subsection{Ground Truth Evaluation}

\setlength{\fboxsep}{0pt}
\begin{figure*}[t!]
\captionsetup[subfigure]{justification=centering}
    \centering
    \begin{subfigure}[t]{.32\linewidth}
        \centering
        \fbox{\includegraphics[width=0.85\textwidth]{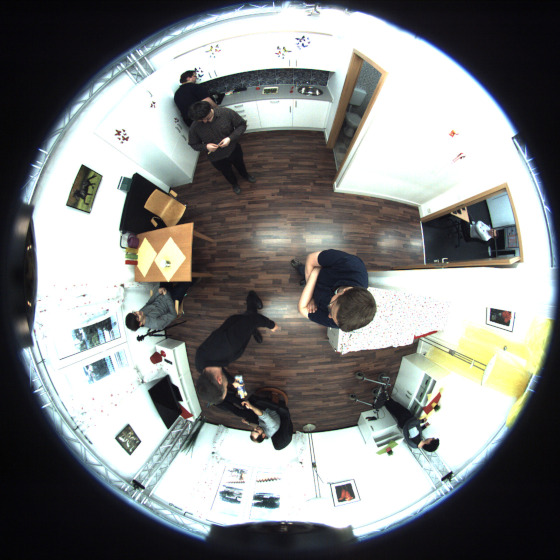}}
        \caption{input data}
        \label{fig:omni_input}
    \end{subfigure}
    \begin{subfigure}[t]{.32\linewidth}
        \centering
        \fbox{\includegraphics[width=0.85\textwidth]{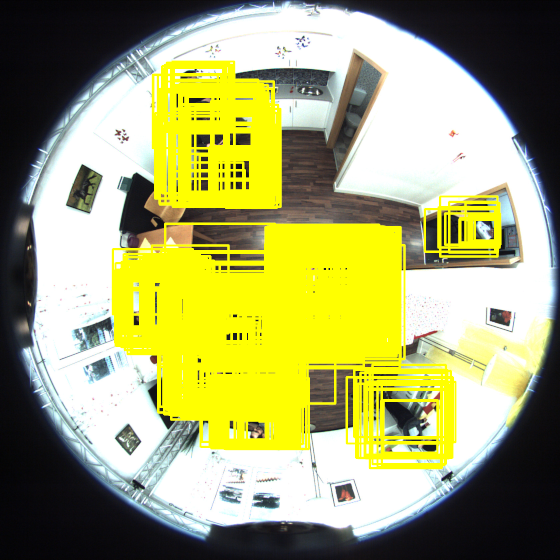}}
        \caption{raw detection boxes}
        \label{fig:raw_yolov2}
    \end{subfigure}
    \begin{subfigure}[t]{.32\linewidth}
        \centering
        \fbox{\includegraphics[width=0.85\textwidth]{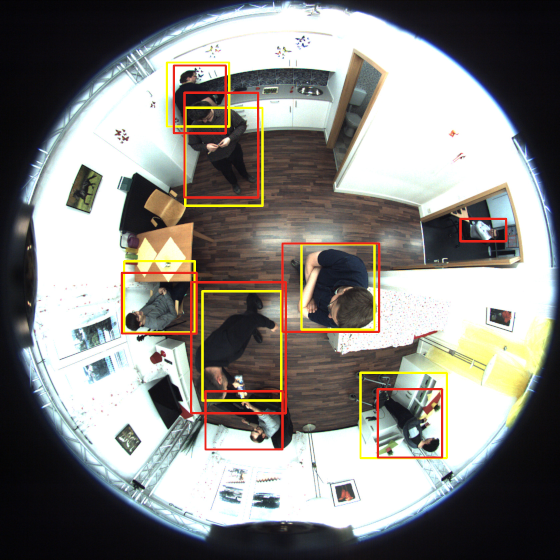}}
        \caption{ground truth (red), detection (yellow)}
        \label{fig:08_01}
    \end{subfigure}
    \caption{Detection result on omnidirectional images for the class person on the Flat dataset.}
    \label{fig:good_bb_gt}
\end{figure*}

% For the qualitative evaluation of our results with respect to the ground truth we use the PIROPO database.
A well working example of our approach is shown in Figure~\ref{fig:good_bb_gt}.
%in the Supplementary Material.
In Figure~\ref{fig:omni_input} we show an omnidirectional input image from our own dataset.
The raw detections of YOLOv2 with a high number of possible true positive candidates without NMS is visualized in Figure~\ref{fig:raw_yolov2}.
The final detection result after the bounding box refinement is shown in Figure~\ref{fig:08_01}.
We apply Soft-NMS with a Gaussian smoothing function.
The ground truth evaluation is done through manually annotated bounding box as shown in Figure~\ref{fig:08_01}.
% We annotate manually a subset of the PIROPO dataset through the consideration of the class person.
% The database is annotated by point-based annotations at the persons center of their heads, which was expanded by us to bounding box ground truth manually.

As scalar evaluation metrics for the detector's result we choose precision and recall \cite{Szeliski2010}, which leads to precision-recall (PR) curves.
Additionally, we determine the AP \cite{Szeliski2010}.
Based on our application we concentrate on the class person, that makes the use of mAP obsolete for evaluation.

The precision and recall are based on the three basic error rates, namely the true positives (TP), the false positives (FP) and the false negatives (FN).
Based on the number of these values per frame in the dataset the precision \(\mathit{pr}\) and recall \(\mathit{re}\) are given by:
\begin{equation}
		\mathit{pr} = \frac{\#TP}{\#TP+\#FP} \text{\quad and \quad} \mathit{re} = \frac{\#TP}{\#TP+\#FN}.
	\label{eq:precision_recall}
\end{equation}

Ideally, the \(\mathit{pr}\) and \(\mathit{re}\) values in  \eqref{eq:precision_recall} are close to one, each.
The higher the values of the evaluation metrics, the larger the area under the PR curve, the better the performance of the detector.

\setlength{\fboxsep}{0pt}
\begin{figure*}[t!]
\centering
    \begin{subfigure}[t]{.48\linewidth}
        \includegraphics[width=\textwidth]{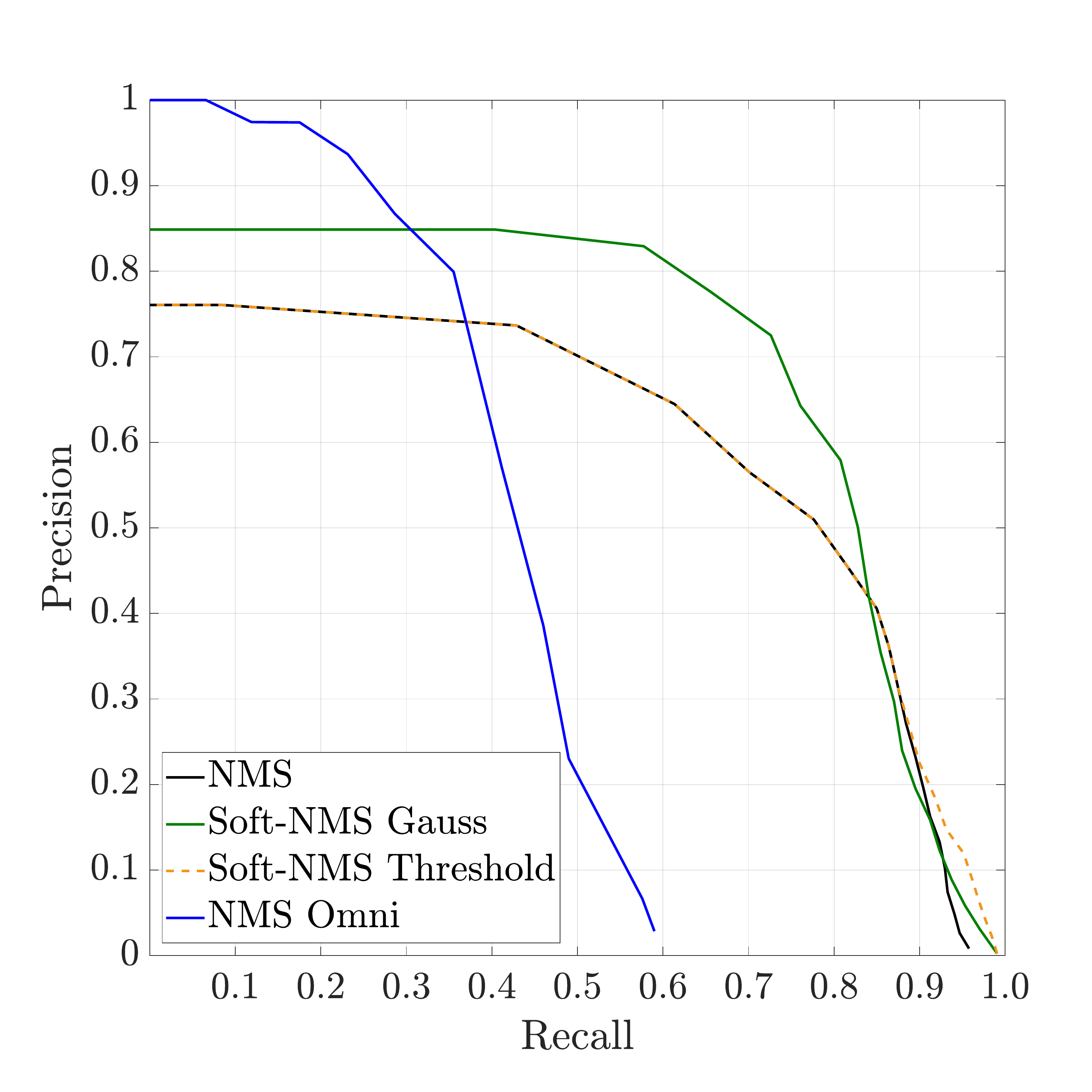}
        \caption{Precision-recall curve on PIROPO}
        \label{fig:0405}
    \end{subfigure}
    \begin{subfigure}[t]{.48\linewidth}
        \includegraphics[width=\textwidth]{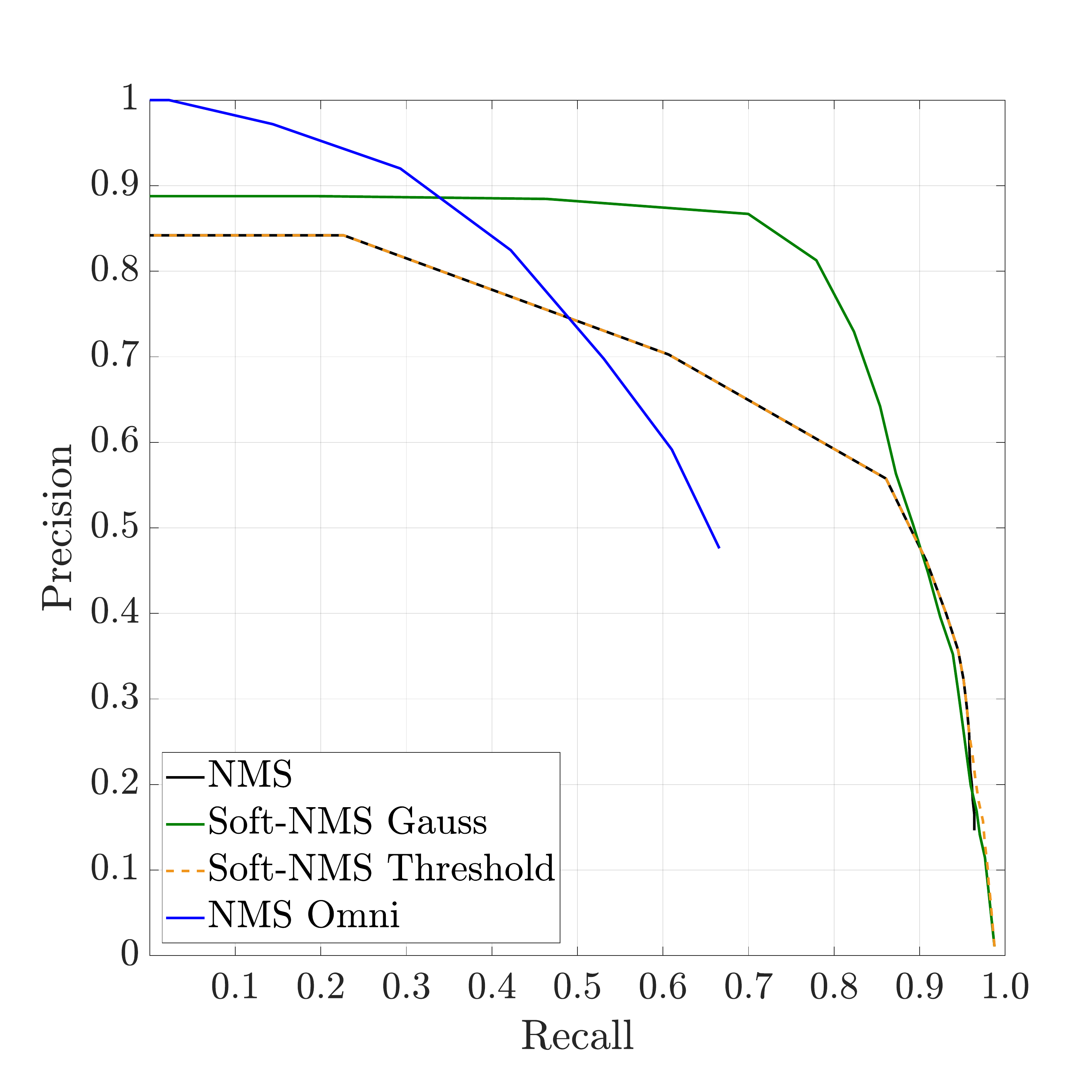}
        \caption{Precision-recall curve on our own data (Flat)}
        \label{fig:0405}
    \end{subfigure}
    \caption{The precision-recall curve for NMS, Soft-NMS and Soft-NMS with Gaussian smoothing function on two different omnidirectional image datasets from back-projected perspective views.
    The precision-recall curve of NMS Omni show the direct application of YOLOv2 to the omnidirectional images.
    \(O_{t} = 0.5\) is the overlap threshold for the IoU from the resulting detection box to the ground truth box.}
    \label{fig:prcurves}
\end{figure*}

The PR curves in Figure~\ref{fig:prcurves} show the evaluation of our method with manually generated ground truth.
The parameter \(O_{t}\) is the overlap threshold for the IoU from the resulting detection box to the ground truth box.
We consider the PR curves for NMS, Soft-NMS and Soft-NMS with a Gaussian smoothing function on omnidirectional images from back-projected perspective views and compare them to the results of the direct application of YOLOv2 to the omnidirectional images, namely NMS Omni.

The steepest curve in Figure~\ref{fig:prcurves} is NMS Omni that reaches a precision of 1 at small recall.
The constellation validates our observations, that the YOLOv2 detector localizes the objects in omnidirectional images accurate with a high number of false negatives.

\begin{table}[ht!]
\caption{Evaluation on PIROPO and Flat dataset with average precision for class person}
\label{tab:apperson}
\centering 
\begin{tabular}{lcc}
\toprule
                 & PIROPO        & Flat \\
\midrule
NMS              & 56.6          & 68.3\\
Soft-NMS Gauss   & \textBF{64.6} & \textBF{77.6} \\
Soft-NMS         & 57.1          & 68.1 \\
Omni             & 41.4          & 69.6 \\
\bottomrule
\end{tabular}
\end{table}

For further quantitative evaluation we compute the AP that is the area under the PR curves of Figure~\ref{fig:prcurves} and visualized in Table~\ref{tab:apperson}. 
The overlap threshold \(O_{t} = 0.5\) follows the PASCAL VOC notation \cite{Everingham2010}.
Additionally, we determine the weighted mean values of precision for NMS, Soft-NMS with Gaussian smoothing, Soft-NMS and apply YOLOv2 to the omnidirectional images directly.
The best (i.e. highest) value of AP is highlighted in bold.
We reach an AP for the class person of \(64.6\%\) through Soft-NMS with a Gaussian smoothing function on PIROPO and \(77.6\%\) on the \(Flat\) Dataset, respectively.

Salient points of the PR curves in Figure~\ref{fig:prcurves} are intersections of the worst performing and the highest performing approach.
Looking at the NMS Omni graph (blue) and the Soft-NMS Gauss graph (blue) we observe an intersection at a precision of \(0.83\) and recall of \(0.35\).
From this point up to recall of \(0.75\) the bounding box refinement method with Soft-NMS Gauss outperforms all other curves without significant decrease of precision on PIROPO.

Another people detector on omnidirectional images is \cite{krams2017} that use DET curves on BOMNI database \cite{bomni2012} for evaluation, therefore we don't compare our results to this people detection approach.
Due to unavailable public training datasets with labeled fish-eye images, we did not do fine-tuning of YOLOv2 from initial weights with omnidirectional image data.

We make the following observations.
After the back projection from the perspective to the omnidirectional view bounding boxes are oversized, because the axis parallelism is not preserved.
Through forcing parallel box edges with respect to the axis in the omnidirectional image coordinate system, we do not receive minimal enclosing rectangles.
% The reason for this is the constancy of changing box size during the  pruning step through NMS and Soft-NMS with Gaussian smoothing (Soft-NMS without a threshold only changes confidences and not prune bounding boxes).
%\item

For the most of the recall and precision values the graphs of NMS and Soft-NMS are equal. Only at precisions smaller than \(0.2\) we observe different trends as shown in Figure~\ref{fig:prcurves}.
%\end{itemize}

\section{Conclusion}
\label{sec:conclu}
\myIndent
In this work we present a method to detect persons in omnidirectional images based on CNNs.
% In previous research activities the synthetic and real-world ground truth generation was well explored.
We apply a state-of-the-art object detector, namely YOLOv2, to virtual perspective views and transform the detections back to the omnidirectional source images. 
% The transformation for covering the azimuth and elevation of the fish-eye image with hundreds of perspective images we call forward mapping.
For the transformation the step size of the two angles, azimuth and elevation was selected in a way, that the perspective images are highly overlapped.
%We detect persons with the current average precision rate of YOLOv2 on perspective images.
In contrast to the standard implementation of YOLOv2 we use the raw detection boxes instead of applying a NMS as bounding box refinement at the end of the network.
After back projection from perspective to omnidirectional images we apply three different NMS methods for pruning the back-projected bounding boxes based on confidence and overlap.

We evaluated the bounding box refinement methods, NMS, Soft-NMS with a threshold and Soft-NMS with Gaussian smoothing on our manually generated ground truth on the PIROPO database and the Flat dataset using PR curves and AP.
At \(O_{t} = 0.5\) we reach an AP for the class person of \(64.6\%\) on PIROPO and \(77.6\%\) on Flat through Soft-NMS with Gaussian smoothing.

%recall, precision and the average precision.
%We reach an improvement of about 41\,\% for the F1-measure compared to directly applying the YOLOv2 to the omnidirectional image data.

%future work

Based on the work of transformation from omnidirectional to perspective and vice versa there are a couple of ideas for future work.
One of our central questions is: how the detection rate of the object detector changes if we consider the lens distortion parameters?

To close the gap of missing omnidirectional ground truth, we will create labeled synthetic and real-world data.
To simplify the data generation we can use our approach followed by manual refinement of detections to create ground truth on omnidirectional images.
% An additional step can be the decision for the projection of the perspective image to a location in the omnidirectional image.
% With both synthetic or real-world data the transformation from perspective to omnidirectional images gives us the possibility of online learning of each arbitrary detector.
%\textcolor{red}{
To improve the approach at the point of projecting bounding boxes from perspective to omnidirectional model it is necessary to minimize the effect of oversized boxes in omnidirectional images.%}
% Various functions in the Soft-NMS step such as generalized logistic functions like the Gompertz function regards the bounding box size.

\section*{Acknowledgements}
This work is funded by the European Regional Development Fund (ERDF) and the Free State of Saxony under the grant number 100-241-945.

%\section*{References}

% References follow the acknowledgments. Use unnumbered first-level
% heading for the references. Any choice of citation style is acceptable
% as long as you are consistent. It is permissible to reduce the font
% size to \verb+small+ (9 point) when listing the references. {\bf
%   Remember that you can use more than eight pages as long as the
%   additional pages contain \emph{only} cited references.}
% \medskip
%\vfill
\bibliographystyle{apalike}
{\small
\bibliography{lit}}

\end{document}